\pdfoutput=1

\documentclass[11pt]{article}

\usepackage[final]{acl}

\usepackage{times}
\usepackage{latexsym}

\usepackage[T1]{fontenc}

\usepackage[utf8]{inputenc}

\usepackage{microtype}

\usepackage{booktabs}
\usepackage{arydshln}
\usepackage{subcaption}
\usepackage{rotating,csquotes}
\usepackage{xcolor}
\usepackage{multirow}
\usepackage{framed}
\usepackage{booktabs}
\usepackage{amsmath}
\usepackage{upgreek}
\usepackage{ulem}
\usepackage{amsfonts}

\usepackage[most]{tcolorbox}

\newtcbox{\mybox}[1][]{enhanced, colframe=blue, colback=blue!15, 
	frame style={opacity=0.25}, interior style={opacity=0.25}, 
	nobeforeafter, tcbox raise base, shrink tight, extrude by=1mm, #1}

\newcommand{\bert}{\textsc{Bert}}

\newcommand{\mightmention}[1]{}
\newcommand{\problem}[1]{\textcolor{red}{$\star$}}
\newcommand{\answer}[1]{\textcolor{blue}{$\#$}}
\newcommand{\todoreview}[1]{\textcolor{green}{$@$}}

\title{RAIL-KD: RAndom Intermediate Layer Mapping for Knowledge Distillation}

\author { 
    Md Akmal Haidar$^1$\thanks{\hspace{2mm}This work has been done while Md Akmal Haidar was at Huawei.} \hspace{3mm}
    Nithin Anchuri$^{1,2}$\thanks{\hspace{2mm}This work has been done while Nithin Anchuri was at Huawei.} \hspace{3mm}
    Mehdi Rezagholizadeh$^1$\hspace{3mm} \\
    \textbf{Abbas Ghaddar$^1$\hspace{3mm}
    Philippe Langlais$^2$ 
    Pascal Poupart$^3$} \\
    $^1$ Huawei Noah's Ark Lab\\
    $^2$ RALI/DIRO, Universit\'e de Montr\'eal, Canada\\
    $^3$ David R. Cheriton School of Computer Science, University of Waterloo \\
    \texttt{\{mehdi.rezagholizadeh,abbas.ghaddar\}@huawei.com}\\ \texttt{felipe@iro.umontreal.ca, ppoupart@uwaterloo.ca}
}
\begin{document}
\maketitle

\begin{abstract}

Intermediate layer knowledge distillation (KD) can improve the standard KD technique (which only targets the output of teacher and student models) especially over large pre-trained language models. However, intermediate layer distillation suffers from excessive computational burdens and engineering efforts required for setting up a proper layer  mapping. To address these problems, we propose a RAndom Intermediate Layer Knowledge Distillation (RAIL-KD) approach in which, intermediate layers from the teacher model are selected randomly to be distilled into the intermediate layers of the student model. This randomized selection enforce that: all teacher layers are taken into account in the training process, while reducing the computational cost of intermediate layer distillation. Also, we show that it act as a regularizer for improving the generalizability of the student model. We perform extensive experiments on GLUE tasks as well as on out-of-domain test sets. We show that our proposed RAIL-KD approach outperforms other state-of-the-art intermediate layer KD methods considerably in both performance and training-time.

\end{abstract}

\begin{table*}[!htp]
    \centering 
    \begin{tabular}{l|ll|l} 
    \toprule
    Model & Layer Mapping  & Complexity &  Limitation \\ 
    \hline
    PKD~\cite{PKD} & Extra Hyperparameter  & $O(m)$  & Extensive Search\\ 
    CKD~\cite{CKD} & Extra Hyperparameter  & $O(m)$  & Extensive Search \\ 
    ALP-KD~\cite{passban2020alp}  & Attention  & $O(m \times n)$  & Slow Training time\\ 
    CoDIR~\cite{CODIR} & Contrastive Learning  & $O(K \times m)$ & Slow Training time\\ 
    \hline 
    RAIL-KD\textsuperscript{$l$} (our) &  Random Selection  & $O(m)$  & -\\ 
    RAIL-KD\textsuperscript{$c$} (our) & Random Selection  & $O(m)$ &-\\ 
    \bottomrule
    \end{tabular}

    \caption{Main Characteristics and limitation of different approaches that tackle the \textit{skip and search problem}. Concat indicates if the approach support concatenated layers distillation. $n$ and $m$ refer to the teacher and student layer number respectively, while $K$ refer to number of negative samples of CODIR. }
    \label{tab:summary} 

\end{table*}

\section{Introduction}
\label{intro}

Pre-trained Language Models (PLMs), such as BERT~\citep{Bert}, RoBERTa~\citep{Roberta} and XLNet~\citep{XLNet} have shown remarkable abilities to match and even surpass human performances on many Natural Languages Understanding (NLU) tasks~\cite{rajpurkar2018know,GLUE,wang2019superglue}. However, the deployment of these models in real world applications (e.g. edge devices) come with challenges, mainly due to large model size and inference time.

In this regard,  several model compression techniques such as quantization~\citep{Q-Bert,Q8-Bert}, pruning~\citep{prune1,prune2,16heads}, optimizing the Transformer architecture~\citep{layerdrop,ghaddar2019contextualized,block2,dynabert}, and knowledge distillation~\citep{Distilbert,Tiny,Mobile,wang2020minilmv2,rashid2021mate,passban2020alp,jafari2021annealing,kamalloo2021not} have been developed to reduce the model size and latency, while maintaining comparable performance to the original model.

KD, which is the main focus of this work, is a neural model compression approach that involves training a small \textit{student} model with the guidance of a large pre-trained \textit{teacher} model. In the original KD technique~\cite{bucilua2006model,KD,turc2019well}, the teacher output predictions are used as soft labels for supervising  the training of the student. There has been several attempts in the literature to reduce the teacher-student performance gap by leveraging   data augmentation~\cite{fu2020role,li2021select,Tiny}, adversarial training~\cite{zaharia2021dialect,rashid2020towards,rashid2021mate}, and intermediate layer distillation (ILD) ~\cite{wang2020minilm,wang2020minilmv2,ji2021show,passban2020alp}.

When it comes to \bert{} compression, ILD leads to clear gains in performances~\cite{Distilbert,Tiny,wang2020minilmv2} due to its ability to enhance the knowledge transfer beyond logits matching. This is done by mapping intermediate layer representations of both models to a common space\footnote{In some cases, the representations are directly matched if the teacher and student have the same hidden size.}, and then matching them via regression~\cite{PKD} or cosine similarity~\cite{Distilbert} losses. On major problem with ILD is the absence of an appropriate strategy to select layers to be matched on both sides, reacting to the \textit{skip and search problem}~\citep{passban2020alp}. There are some solutions in the literature mostly rely on layer combination~\citep{CKD}, attention-based layer projection~\citep{passban2020alp} and contrastive learning~\citep{CODIR}.

While these solutions are all effective to some extent, to the best of our knowledge, there is no work in the literature doing a comprehensive evaluation of these techniques in terms of both efficiency and performance.  A case in point is that the aforementioned solutions to the layer skip and search problem are not scalable to very deep networks. We propose RAIL-KD (RAndom Intermediate Layer KD), a simple yet effective method for intermediate layer mapping which randomly selects $k$ out of $n$ intermediate layers of the teacher at each epoch to be distilled to the corresponding student layers. Since the layer selection is done randomly, all the intermediate layers of the teacher will have a chance to be selected for distillation. Our method adds no computational cost to the training, still outperforming all aforementioned methods on the GLUE benchmark~\cite{GLUE}. Moreover, we observe larger gains distilling larger teacher models, as well as when compressed models are evaluated on out-of-domain datasets. Last, we report the results on 5 random seeds in order to verify the contribution of the random selection process, thus making the comparison fair with previous methods. The main contributions of our paper are as follows:

\begin{itemize}
    \item We introduce RAIL-KD, a more efficient and scalable intermediate layer distillation approach. 
    \item To the best of our knowledge, we are the first to perform a comprehensive study of the intermediate layer distillation techniques in terms of both efficiency and performance. 
    \item We consider the distillation of models such as \bert{} and RoBERTa, and compare different up-to-date distillation techniques on out-of-domain test sets. Thus providing new points of comparison.
\end{itemize}

\begin{figure*}[!htb]
    \centering
    \includegraphics[width=15cm]{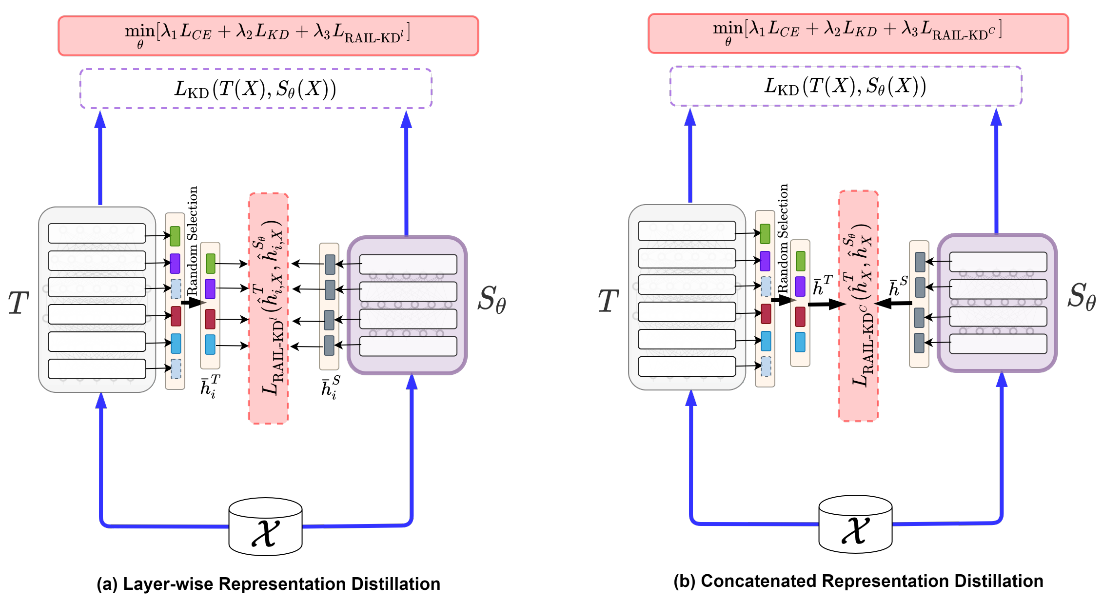}
    \caption{Proposed RAIL-KD technique for efficient intermediate layer distillation. (a) This version shows a layer-wise projection which is indicated as RAIL-KD$^l$ in the paper. (b) This variant named RAIL-KD$^c$, concatenates the intermediate representations of each network before distillation.  }
    \label{fig:IRD}
\end{figure*}

\section{Related Work}

Recent years, have seen a wide range of methods have emerged aiming to expand knowledge transfer of transformer-based~\cite{vaswani2017attention} NLU models beyond logits matching. Distill\bert{}~\cite{Distilbert} added a cosine similarity loss between teacher and student embeddings layer. TinyBERT~\cite{Tiny}, MobileBERT~\cite{Mobile}, and MiniLM~\cite{wang2020minilm} matched the intermediate layers representations and self-attention distributions of the teacher and the student.

In PKD, \citet{PKD} used deterministic mapping strategies to distill a 12-layers \bert{} teacher to a 6-layers \bert{} student. \textsc{PKD-Last} and \textsc{PKD-Skip} refer to matching layers $\{1-5\}$ of the student with layers $\{7-11\}$ and $\{2,4,6,8,10\}$ of the teacher respectively. 
However, these works ignored the impact of layer selection, as they used a fixed layer-wise mapping.\footnote{e.g. matching the first (or last) $k$ layers of the student with their corresponding teacher layers.}

Researchers have found that tuning the layer mapping scheme can significantly improve the performance of ILD techniques~\cite{PKD}. Nevertheless, finding the optimal mapping can be challenging, which is referred to as the \textit{layer skip and search problems} by ~\cite{passban2020alp}. 
To address the layer skip problem, CKD~\citep{CKD} is built on top of PKD by partitioning all the intermediate layers of the teacher to the number of student layers. Then, the combined representation of the layers of each partition is distilled into a number of subset corresponding to the number of student layers. However, finding the optimal partitioning scheme requires running exhaustive experiments. 

Given teacher and student \bert{} models with $n$ and $m$ layers  respectively (where $n>>m$), it is not trivial to choose the teacher layers that can be incorporated in the distillation process and  how we should map them to the student layers (\textit{search}).

ALP-KD~\cite{passban2020alp} overcomes this issue by computing attention weights between each student layer and all the intermediate layers of the teacher. 
The learned attention weights for each student layer are used to obtain a weighted representation of all teacher layers. Although ALP-KD has shown promising results on 12-layer BERT-based compression, attending to all layers of the teacher adds considerable computational overhead to the training phase. This can become computationally prohibitive when scaling to very large models such as RoBERTa-large~\citep{Roberta} or GPT-2~\citep{GPT}. 
Alternatively, CODIR~\cite{CODIR} exploited contrastive learning~\citep{CRD} to perform intermediate layers matching between the teacher and the student models with no deterministic mapping. Similar to ALP-KD, this approach also requires excessive training time due to the contrastive loss calculation and the use of negative samples from a memory bank.

Table~\ref{tab:summary} summarizes the main characteristics of the existing state-of-the-art intermediate layer distillation techniques used for pre-trained language models compared with our proposed RAIL-KD. As it is shown in this table, PKD~\cite{PKD}, CKD~\cite{CKD}, and CoDIR are the most related works to us. However, PKD and CKD treat the mapping as an extra hyperparameter that requires extensive experiments to find the optimal mapping. On the other hand, ALP-KD~\cite{passban2020alp} and CoDIR~\cite{CODIR} use attention mechanism and contrastive learning respectively to address the issue, but at the expense of extra computational cost. 

Our proposed RAIL-KD method does not add any computational cost to the distillation process, while empirically outperforming previous methods. For instance, RAIL-KD is roughly twice faster than CoDIR in a 24 to 6 layers compression. In addition, it does not require extensive experiments to find the optimal mapping scheme. In this work, we position ourselves to works that tackle the \textit{skip and search problem}~\footnote{Only work that performs intermediate layer distillation}. Otherwise said, we don't compare with works like TinyBERT~\cite{Tiny} or MiniLM~\cite{wang2020minilm}, which use extra losses like self-attention distribution matching. However, we expect that these methods, as well as state-of-the-art~\cite{rashid2021mate,he2021generate} one can take full advantage of RAIL-KD, since they use deterministic layer mapping scheme.

\section{RAIL-KD}

The RAIL-KD method is sketched in Figure~\ref{fig:IRD}. In contrast to traditional intermediate layer distillation techniques which keep the selected layers of the teacher for distillation fixed during training, in RAIL-KD, at each epoch, a few intermediate layers from the teacher model are selected randomly for distillation. Here for simplicity, we set the number of selected intermediate layers of the teacher model equal to that of the student model. Our method is architecture agnostic can be applied to different domains such as computer vision and NLP.

RAIL-KD transfers intermediate knowledge of a pre-trained teacher $T$ with $n$ intermediate layers to a student model $S_\theta$ with $m$ intermediate layers. Let $(X,y)$ denotes a training sample $X=(x_0,\cdots,x_{L-1})$  which is a sequence of $L$ (sub-)tokens and $y$ its corresponding label. 
In Figure~\ref{fig:IRD}, our Random Selection operator is applied to the intermediate layers of the teacher to randomly select $m$ out of $n$ layers. 
The intermediate layer representations of the $m$ selected layers of the teacher and the student model corresponding to the $X$ input can be described as $H_X^T=\{H_{1,X}^T,\cdots,H_{m,X}^T\}$ and $H_X^{S_\theta}=\{H_{1,X}^{S_\theta},\cdots,H_{m,X}^{S_\theta}\}$ respectively, where $H_{i,X}^T = \cup_{k=0}^{L-1} \{H_{i,x_k}^T\} \in R^{L\times d_1}$ and $H_{i,X}^{S_\theta}= \cup_{k=0}^{L-1} \{H_{i,x_k}^{S_{\theta}}\} \in R^{L\times d_2}$.

Here, $d_1$ and $d_2$ indicate the hidden dimension of the layers of the teacher and the student models respectively. To obtain $H_{i,X}^T$ and $H_{j,X}^S$, we need to find an aggregated representation for the sequence of $L$ tokens at each layer of the two networks. In this regard, one can either use the $<$CLS$>$ token representation or use the mean-pooling of the sequence representations of the layer.
Since in~\citep{CODIR}, the mean-pooling representation shows better results, we adopt it to compute the sentence representation of each layer.  Mean-pooling is a row-wise average over 
$H_{i,X}^T$, $H_{i,X}^{S_\theta}$ to get $\bar{h}_{i,X}^T  \in  R^{d_1}$
$\bar{h}_{i,X}^{S_\theta} \in R^{d_2}$ ~\citep{CODIR}:

\begin{equation}
    \begin{split}
         \bar{h}_{i,X}^T = \frac{1}{L} \sum_{k=0}^{L-1} H_{i,x_k}^T \hspace{2mm};\hspace{2mm}
         \bar{h}_{i,X}^{S_\theta} = \frac{1}{L} \sum_{k=0}^{L-1} H_{i,x_k}^{S_\theta} 
    \end{split}
\end{equation}
RAIL-KD proposes the intermediate layer  distillation in two different forms: using layer-wise distillation (see Fig. 1(a)) or by concatenating layer representations (see Fig. 1(b)). 

\subsection{Layer-wise RAIL-KD}
In this setting, the representations 
$\bar{h}_{i,X}^T  \in  R^{d_1}$ and 
$\bar{h}_{i,X}^{S_\theta} \in R^{d_2}$ are projected into the same lower-dimensional space 
$\hat{h}_{i,X}^T,\hat{h}_{i,X}^{S_\theta} \in R^u$ using $(d_1 \times u)$ and $(d_2 \times u)$ linear mappings respectively to calculate the layer-wise losses. 
\begin{equation}
\begin{split}
& L_{\text{RAIL-KD}^l}=  \sum_{X\in \mathcal{X}} \sum_{i=1}^m \alpha_i \left( \lvert\lvert  \frac{\hat{h}_{i,X}^T}{\lvert\lvert\hat{h}_{i,X}^T \rvert\rvert_2}-\frac{\hat{h}_{i,X}^{S_\theta}}{\lvert\lvert\hat{h}_{i,X}^{S_\theta} \rvert\rvert_2}\rvert\rvert_2^2  \right)    
\end{split}
\label{eq1}
\end{equation}
where $\mathcal{X}$ denotes the set of training samples, and $\alpha_i$ is a hyper-parameter to weigh the layer-wise distillation loss.   

\subsection{Concatenated RAIL-KD}

In this setting, intermediate layer representations are concatenated and then distilled:  
$\Bar{h}_X^T=[\Bar{h}_{1,X}^T,\cdots,\Bar{h}_{m,X}^T]$,
$\Bar{h}_X^{S_\theta}=[\Bar{h}_{1,X}^{S_\theta},\cdots,\Bar{h}_{m,X}^{S_\theta}]$ which are further mapped into the same lower-dimensional space 
$\hat{h}_X^T,\hat{h}_X^{S_\theta} \in R^u$ using $(md_1 \times u)$ and $(md_2 \times u)$ linear mappings to calculate the concatenated distillation loss. 
\begin{equation}
L_{\text{RAIL-KD}^c}=\sum_{X\in \mathcal{X}} \lvert\lvert  \frac{\hat{h}_X^T}{\lvert\lvert\hat{h}_X^T \rvert\rvert_2}-\frac{\hat{h}_X^{S_\theta}}{\lvert\lvert\hat{h}_X^{S_\theta} \rvert\rvert_2}\rvert\rvert_2^2 
\label{eq2}
\end{equation}
Any type of loss such as  contrastive~\citep{CODIR}, or mean-square-error (MSE)~\citep{passban2020alp,PKD} can be applied for our RAIL-KD approach.

\subsection{Training Loss}
The intermediate representation distillation loss $L_{\text{RAIL-KD}}$ is combined with the original KD loss $L_{\text{KD}}$, which is used to distill the knowledge from the output logits of the teacher model $T$ to the output logits of the student model $S_\theta$, and the original cross-entropy loss $L_{\text{CE}}$. 
The total loss function for training the student model is:
\begin{equation}
    \mathcal{L}=\lambda_1 L_{\text{CE}} + \lambda_2 L_{\text{KD}} + \lambda_3 L_{\text{RAIL-KD}^{l/c}}
\end{equation}
where $\lambda_1$, $\lambda_2$, and $\lambda_3$ are hyper-parameters of our model to minimize the total loss, and $\lambda_1+\lambda_2+\lambda_3=1$. 

\begin{table*}[!htb]
\centering
\begin{tabular}{lccccccccc}
\toprule
\textbf{Model} & \textbf{CoLA}& \textbf{SST-2}& \textbf{MRPC}& \textbf{STS-B}& \textbf{QQP}& \textbf{MNLI}& \textbf{QNLI}& \textbf{RTE}& \textbf{Avg.}\\

\midrule
\multicolumn{10}{c}{\textsc{Dev}}\\
\midrule

Teacher & 61.3 & 93.0 & 90.6 & 88.4 & 91.0 & 84.7 & 91.5 & 68.2 & 83.7 \\ 
\hdashline
w/o KD & 53.3 & 90.1 & 90.0 & 86.5 & 90.4 & 82.3 & 89.1 & 61.7 & 80.4 \\ 
Vanilla KD & 55.8 & 90.3 & 90.3 & 86.6 & 90.5 & 82.7 & 89.6 & 68.5 & 81.9 \\ 

PKD & 56.1 & 91.3 & 90.7 & 87.4 & 91.2 & 83.3 & 90.2 & 69.3 & 82.5\\ 
ALP-KD & 56.8 & 90.8 & 90.6 & 87.5 & 91.0 & 83.4 & 90.2 & 70.4 & 82.7 \\ 

\hdashline
RAIL-KD$^l$ & \textbf{58.8} & \textbf{92.8} & \textbf{91.0} & 87.8 & 91.2 & 83.5 & \textbf{90.3} & 70.4 & 83.2 \\ 
RAIL-KD$^c$& 57.2 & 91.9 & 90.8 & \textbf{87.9} & \textbf{91.4} & \textbf{83.5} & 90.1 & \textbf{72.2} & \textbf{83.2} \\ 

\midrule
\multicolumn{10}{c}{\textsc{Test}}\\
\midrule

Teacher & 52.0 & 92.9 & 87.8 & 82.3 & 88.9 & 84.3 & 90.7 & 66.0 & 81.0 \\

\hdashline
w/o KD & 50.7 & 91.7 & 87.2 & 80.4 & 88.3 & 81.4 & 88.4 & 57.6 & 78.6 \\ 
Vanilla KD & 50.9 & 91.0 & 87.7 & 81.0 & 88.5 & 82.2 & 88.7 & 60.6 & 79.2\\ 
PKD & 50.6 & 92.0 & 87.2 & 81.7 & 89.1 & 82.7 & 89.0 & 60.6 & 79.5 \\ 
ALP-KD & 50.2 & 90.8 & 87.6 & 81.9 & 89.0 & 82.7 & 88.9 & \textbf{61.8} & 79.5 \\
\hdashline

RAIL-KD$^l$& \textbf{51.3} & 92.3 & 87.9 & \textbf{82.1} & \textbf{89.2} & 82.6 & 89.0 & 60.8& 79.7 \\ 
RAIL-KD$^c$ & 50.6 & \textbf{92.5} & \textbf{88.2} & 81.4 & 88.9 & \textbf{82.8} & \textbf{89.3} & 61.3 & \textbf{79.8} \\ 

\bottomrule

\end{tabular}
\caption{\textsc{Dev} and \textsc{Test} performances on GLUE benchmark when \bert{$_{12}$} and Distill\bert{$_6$} are used as backbone for the teacher and students variants respectively. Bold mark describes the best results.}
\label{tab:bert_glue}
\end{table*}

\section{Experimental Protocol}

\subsection{Datasets and Evaluation}

We evaluate RAIL-KD on 8 tasks from the GLUE benchmark~\cite{GLUE}: 2 single-sentence (CoLA and SST-2) and 5 sentence-pair (MRPC, RTE, QQP, QNLI, and MNLI) classification tasks, and 1 regression task (STS-B). Following prior works~\citep{PKD,passban2020alp,Tiny,CODIR}, we use the same metrics as the GLUE benchmark for evaluation. Moreover, to further show the generalization capability of our RAIL-KD method on out-of-domain (OOD) across tasks, we use Scitail~\citep{SciTail}, PAWS (Paraphrase Adversaries from Word Scrambling)~\citep{PAWS}, and IMDb (Internet Movie Database)~\citep{IMDB}  test sets to evaluate the models fine-tuned on MNLI, QQP, and SST-2 tasks respectively.

\begin{table*}[!htb]
\centering
\begin{tabular}{lccccccccc}

\toprule
\textbf{Model} & \textbf{CoLA}& \textbf{SST-2}& \textbf{MRPC}& \textbf{STS-B} & \textbf{QQP}& \textbf{MNLI}& \textbf{QNLI}& \textbf{RTE} & \textbf{Avg.}\\ 

\midrule
\multicolumn{10}{c}{\textsc{Dev}}\\
\midrule

Teacher & 68.1 & 96.4 & 91.9 & 92.3 & 91.5 & 90.1 & 94.6 & 86.3 & 88.9 \\ 
\hdashline
w/o KD & 56.6 & 93.1 & 89.5 & 87.2 & 91.0 & 84.6 & 91.3 & 65.7 &  82.4 \\ 
Vanilla KD & 60.9 & 93.1 & 90.2 & 89.0 & 91.1 & 84.7 & 91.3 & 68.2 & 83.6 \\ 
PKD & 62.3 & 91.6 & 90.9 & 88.9 & 91.6 & 84.4 & 91.1 & 71.1 & 84.0 \\ 
ALP-KD & 62.7 & 91.7 & 91.1 & 89.1 & 91.4 & 84.3 & 90.8 & 71.1 & 84.0 \\
\hdashline
RAIL-KD$^l$ & \textbf{65.4} & \textbf{93.8} & 90.1 & 89.4 & 91.9 & 84.8 & 92.0 & 72.9 & 85.1 \\ 
RAIL-KD$^c$&65.3 & 93.7 & \textbf{91.4} & \textbf{89.4} & \textbf{92.0} & \textbf{84.8} & \textbf{92.0} & \textbf{72.9} & \textbf{85.2} \\ 


\midrule
\multicolumn{10}{c}{\textsc{Test}}\\
\midrule
Teacher & 68.1 & 96.4 & 91.9 & 92.3 & 91.5 & 90.2 & 94.6 & 86.3 & 85.3 \\
\hdashline
PKD  & 50.2 & 89.4 & 88.9 & 84.5 & 92.3 & 84.0 & 90.2 & 62.7 & 80.3 \\ 
ALP-KD &  53.6 & 89.6 & 89.2 & 84.6 & 92.8 & 83.6 & 90.4 & \textbf{64.4} & 81.0 \\ 
\hdashline

RAIL-KD$^l$ & 53.4 & 89.5 & 88.9 & 84.8 & \textbf{93.6} & \textbf{84.5} & 91.1 & 63.5 & 81.2 \\ 
RAIL-KD$^c$& \textbf{53.6} & \textbf{89.6} & \textbf{89.6} & \textbf{84.8} & 93.4 & 83.9 & \textbf{91.6} & 63.8 & \textbf{81.3} \\ 

\bottomrule

\end{tabular}
\caption{\textsc{Dev} and \textsc{Test} performances on GLUE benchmark when RoBERTa$_{24}$ and DistillRoberta$_6$ are used as backbone for the teacher and student variants respectively. Bold mark describes the best results.}
\label{tab:roberta_glue}
\end{table*}

\subsection{Implementation Details}

We run extensive experiments on 3 different teachers in order to ensure a fair comparison with of a wide range of prior works, and also to show the effectiveness of RAIL-KD. We experiment with the 12 layers \bert{-base-uncased}~\cite{Bert}  as teacher (\bert{$_{12}$}) and 6 layer DistilBert~\cite{Distilbert} as student (Distill\bert{$_6$}) to compare with PKD~\cite{PKD} and ALP-KD~\cite{passban2020alp}. Also, we use  24 layers RoBERTa-large~\citep{Roberta}  and 6 layers DistilRoberta~\citep{DistilRoberta} as the backbone for teacher (RoBERTa$_{24}$) and student (DistilRoberta$_6$) respectively to compare models when  $n >> m$. Furthermore, we perform evaluation using 12 layers RoBERTa-base (RoBERTa$_{12}$) as teacher to be able to directly compare our numbers with the ones of CoDIR.

We re-implement PKD~\citep{PKD} and ALP-KD~\citep{passban2020alp} approaches using the default settings proposed in the respective papers. We used early stopping based on performance on the development set, while making sure that the figures are in line with the ones reported in the papers. More precisely, the best layer setting for PKD teacher BERT\textsubscript{12} is $\{2, 4, 6, 8, 10\}$ to distill into DistilBERT\textsubscript{6}. For DistilRoBERTa\textsubscript{6}, we choose the intermediate layers {4, 8, 12, 16, 20} from the teacher RoBERTa\textsubscript{24} model for distillation that we found to work the best on the development set.

Using ALP-KD, we compute attention weights for the intermediate layers of the teacher (i.e., 1 to 11 for BERT\textsubscript{12}
and 1 to 23 for RoBERTa\textsubscript{24} models)
to calculate the weighted intermediate representations of the teacher for each intermediate layer of the student model (i.e., 1 to 5 layers of the student models). Since, the hidden dimensions of the RoBERTa\textsubscript{24} and DistilRoBERTa\textsubscript{6} are different, we linearly transform them into same lower-dimensional space. We train the PKD and ALP-KD models following~\citep{PKD,passban2020alp}.

For RAIL-KD$^l$, at each epoch we randomly select 5 layers from the intermediate layers of the teacher (i.e., from layers 1 to 11 for BERT\textsubscript{12} model and 1 to 23 for  RoBERTa\textsubscript{24} model). Then, we sort the layer indexes and perform layer-wise distillation (Figure~\ref{fig:IRD}(a)) for RAIL-KD$^l$. For RAIL-KD$^c$, we concatenated the representations of the sorted randomly selected intermediate layers and then perform concatenated representation distillation (Figure~\ref{fig:IRD}(b)). We use a linear transformation to map the intermediate representations (layer-wise or concatenated representations) into 128-dimensional space and normalize them before computing the loss $L_{\text{RAIL-KD}^{l/c}}$ for both BERT\textsubscript{12} and RoBERTa\textsubscript{24} distillations. We fixed $\alpha_i=1$, $\lambda_1, \lambda_2, \lambda_3 = 1/3$ for our proposed approaches~\footnote{We didn't find a significant improvement when changing these values.}. We search learning rate from \{1e-5, 2e-5, 5e-5, 4e-6\}, batch size from \{8, 16, 32\}, and fixed the epoch number to 40 for all the experiments. we run all experiments 5 times and report average score, in order to validate the credibility of our results. We ran all the experiments on a single NVIDIA V100 GPU using mixed-precision training~\citep{fp16} and PyTorch~\cite{paszke2019pytorch} framework.

Our results indicate that random layer mapping not only delivers consistently better results than the deterministic mapping techniques such as PKD, but it has less computational overhead during training, while avoid extensive search experiments to find optimal mapping. On the other hand, using attention for layer selection (ALP-KD) or contrastive learning (CoDIR) leads to slightly worse result than than random selection.

\begin{table*}[!ht]
\centering
\begin{tabular}{l|cccccccc|c}
\toprule
\textbf{Model} & \textbf{CoLA}& \textbf{SST-2}& \textbf{MRPC}&
\textbf{QQP}& \textbf{MNLI}& \textbf{QNLI}& \textbf{RTE}& \textbf{Avg.} & \textbf{Speedup}\\
\midrule
RoBERTa\textsubscript{12} &62.0&95.3&90.1&89.4&87.2&93.2&72.7&84.6&1x \\ 
\midrule
CoDIR&53.6&93.6&\textbf{89.4}&89.1&83.2&90.4&\textbf{65.6}&81.0 &1x\\ 
RAIL-KD\textsuperscript{$c$}&\textbf{54.2}&\textbf{93.6}&88.4&\textbf{89.5}&\textbf{83.9}&\textbf{91.7}&64.5&\textbf{81.2}&\textbf{1.96x} \\ 
\bottomrule
\end{tabular}
\caption{GLUE test results of RAIL-KD$^c$ when using Roberta$_{12}$ and DistilRoBERTa$_6$ as backbone for teacher and students. Results of CoDIR are directly copied from their paper~\cite{CODIR}.}
\label{tab:codir_glue}
\end{table*}

\section{Results}

Table~\ref{tab:bert_glue} shows the performances of models trained on GLUE tasks, and evaluated on their respective \textsc{dev} and \textsc{test} sets for 12 layer to 6 layer distillation. \bert{$^{12}$} and DistilBERT{$^6$} are used as backbone for the teacher and student models respectively. The baselines are fine-tuning without KD (w/o KD) and Vanilla KD. Moreover, we directly compare RAIL-KD$^{lc}$ results with   PKD and ALP-KD as more competitor techniques. 

First, we observe that in the 12 to 6 layer distillation, the performance gap between ILD methods and vanilla-KD is tight (0.8\% and 0.3\% on \textsc{dev} and \textsc{test} sets respectively). Moreover, as we expect, ALP-KD performs better (on \textsc{dev}) and similar (on \textsc{test}) compared to PKD with 0.2\% improvement on the \textsc{dev} results. Second, results show that RAIL-KD outperforms the best ILD methods by margin of 0.5\% and 0.3\% on average on \textsc{dev} and \textsc{test} sets respectively. We notice that, except on RTE \textsc{test}, our RAIL-KD$^{l/c}$ obtained the highest per-task performances. Third, we observe that RAIL-KD$^{l/c}$ perform very similarly, which indicates that our method is effective on concatenated as well as layer-wise  distillation.

Similar trends are seen on the 24 to 6 layer model compression experiments, which are reported in Table~\ref{tab:roberta_glue}. In this experiment, we used Roberta$_{24}$ and DistillRoberta$_6$ as teacher and students models respectively. Overall, RAIL-KD outperforms the best baseline by 1.2\% and 0.3\% on \textsc{dev} and \textsc{test} sets respectively. Interestingly, the gap on \textsc{dev} compared with PKD and ALP-KD is larger than the one reported on \bert{$^{12}$} experiments, and PKD \textsc{Test} socres are much lower from that of ALP and RAIL-KD. This might be because PKD skips a large number of intermediate layers on RoBERTa$_{24}$, and the computational cost of ALP-KD attention weights over a large number of teacher layers might produce smaller weights on Roberta$_{24}$ compared to \bert{$_{12}$}.

Furthermore, we demonstrate the effectiveness of RAIL-KD by directly comparing it with CoDIR~\cite{CODIR}, the current state-of-the-art ILD method. It uses the contrastive objective and a memory bank to extract a large number of negative samples for contrastive loss calculations. Table~\ref{tab:codir_glue} shows GLUE test results of both approaches when distilling RoBERTa$_{12}$ to DistillRoberta$_6$. CoDIR results are adopted from their paper, and we followed their experimental protocol by not reporting scores on \textsc{STS-B}. In addition, we report the overall training time speedup against the teacher for different techniques. On average, RAIL-KD is almost twice faster while performs on par with CoDIR (+0.2\%). Moreover, RAIL-KD outperforms CoDIR on 5 out of 8 datasets.

\begin{figure*}[!thp]
\centering
\includegraphics[height=3.5cm,width=4.31cm]{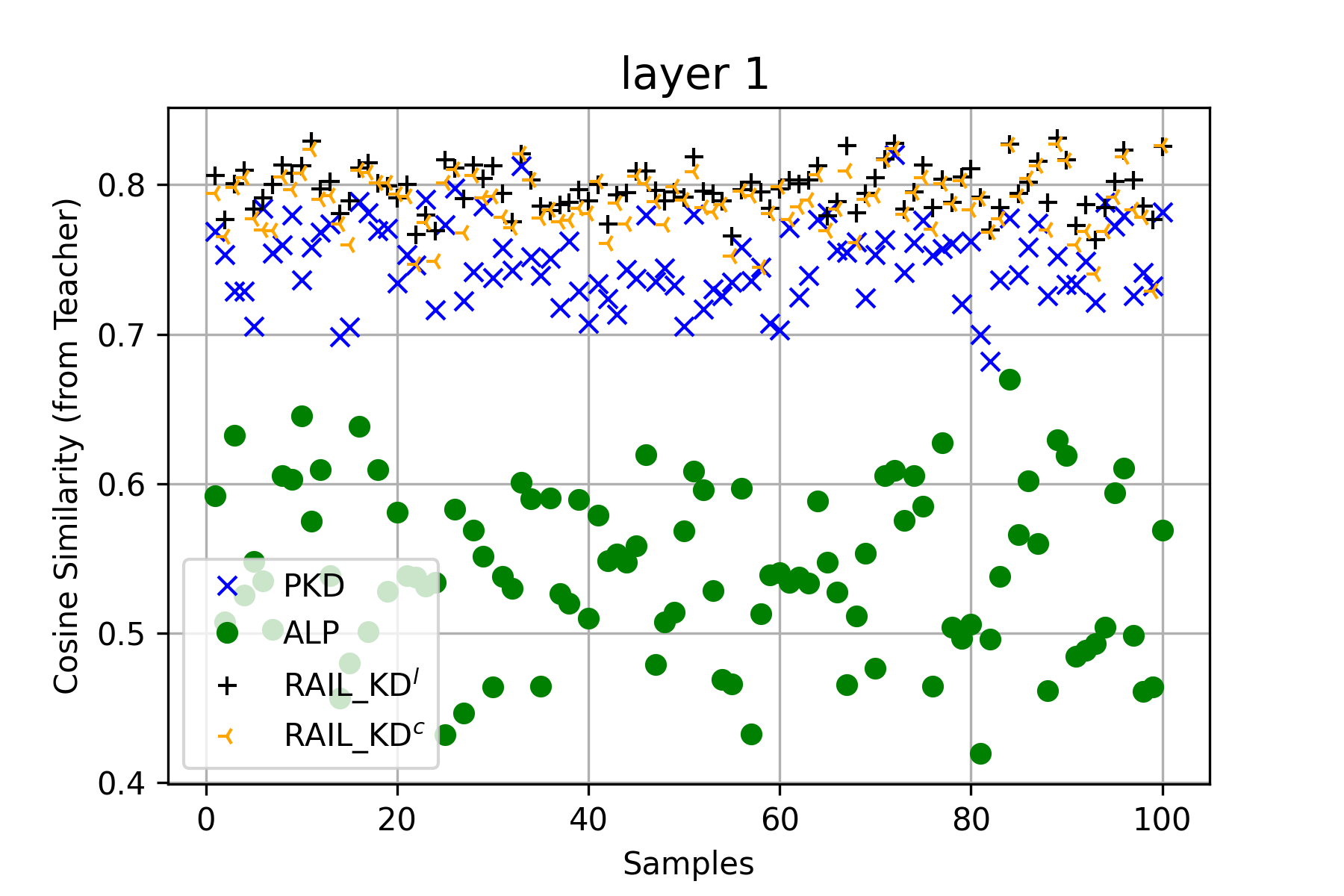}
\includegraphics[height=3.5cm,width=4.31cm]{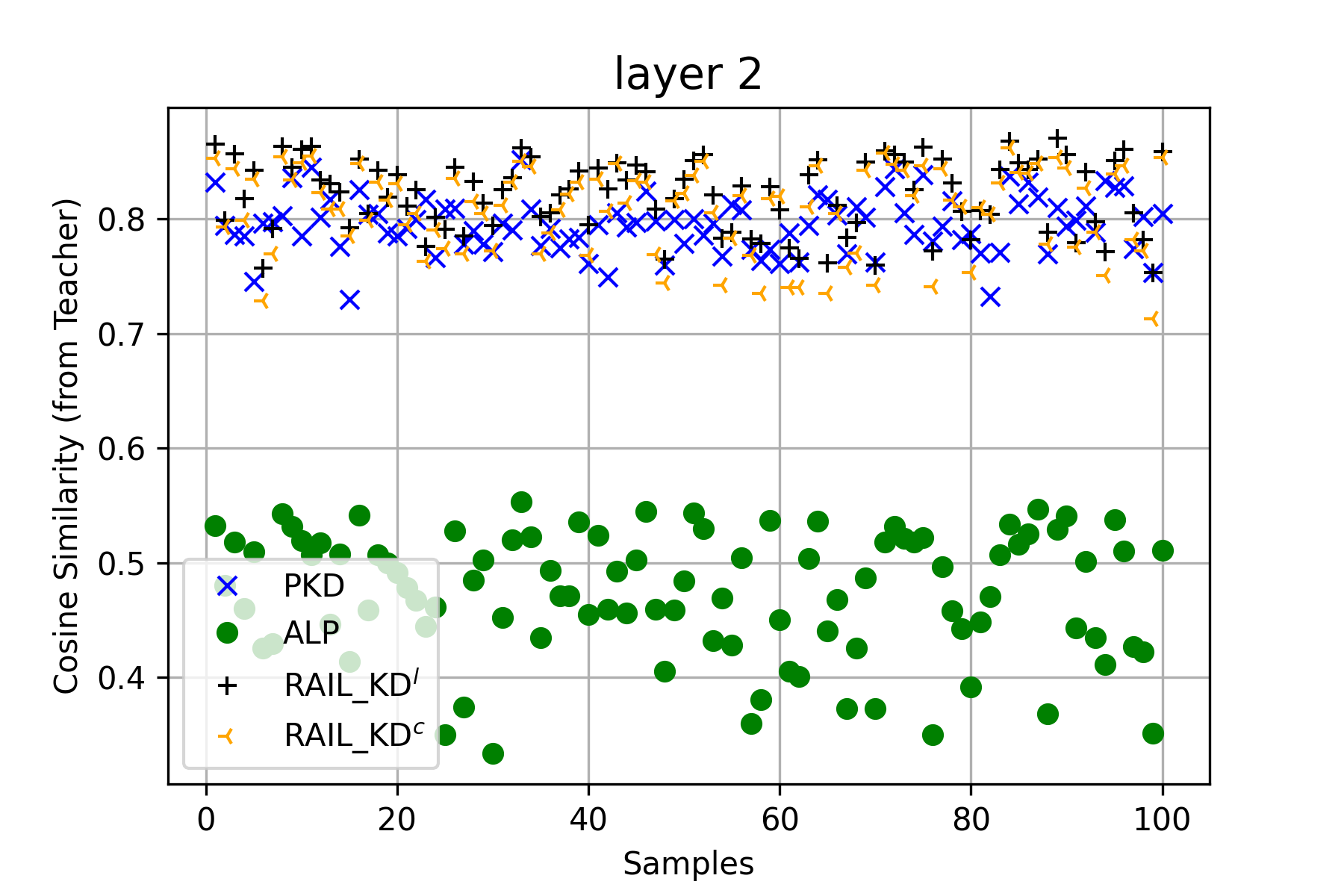}
\includegraphics[height=3.5cm,width=4.31cm]{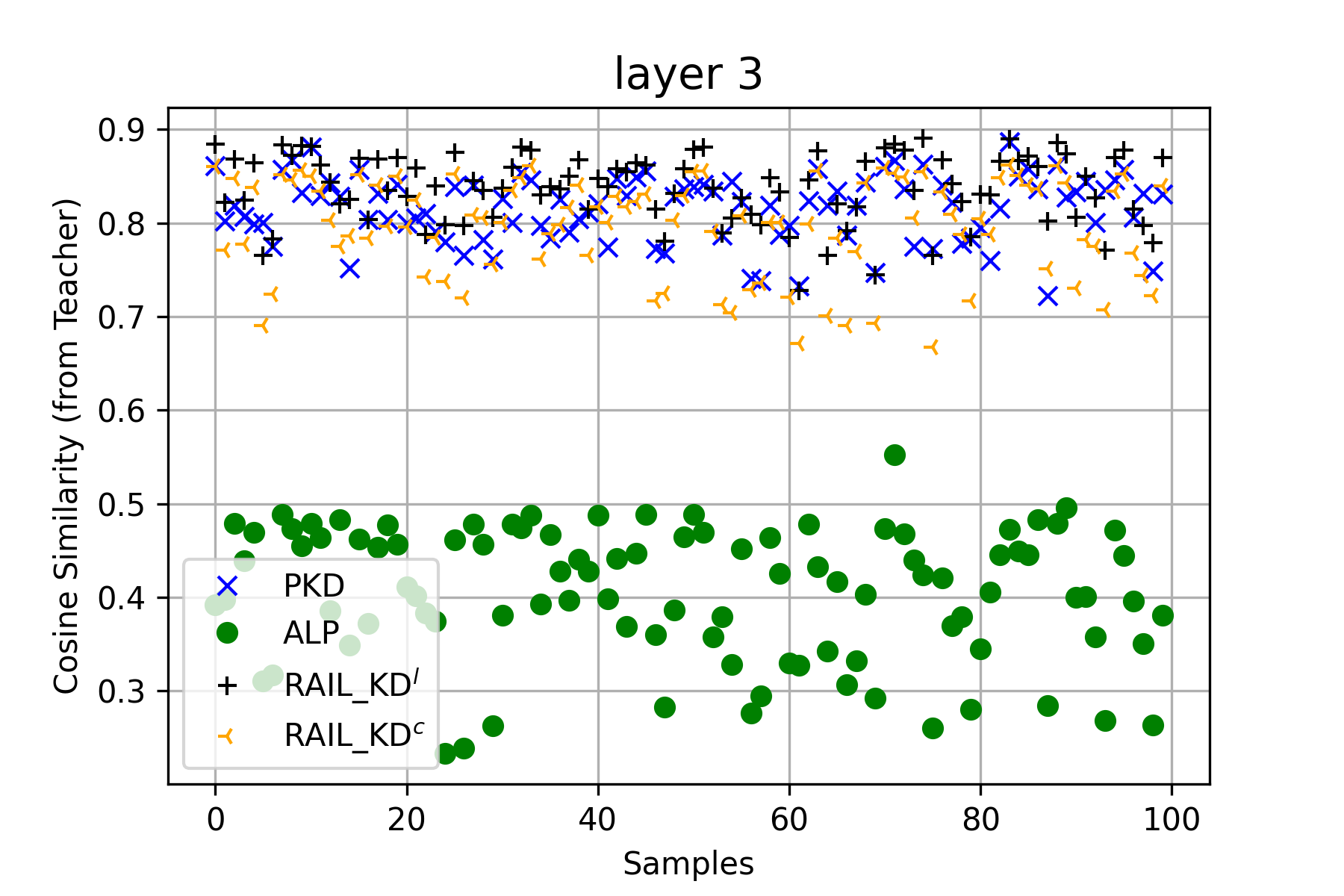}
\caption{Cosine similarity between the intermediate layer representations of the \bert{$_{12}$} teacher and Distill\bert{$_6$} student models compute don the SST-2 dataset.}
\label{fig:plot}

\end{figure*}

\begin{figure*}[!tbp]
\centering
\includegraphics[height=3.5cm,width=4.31cm]{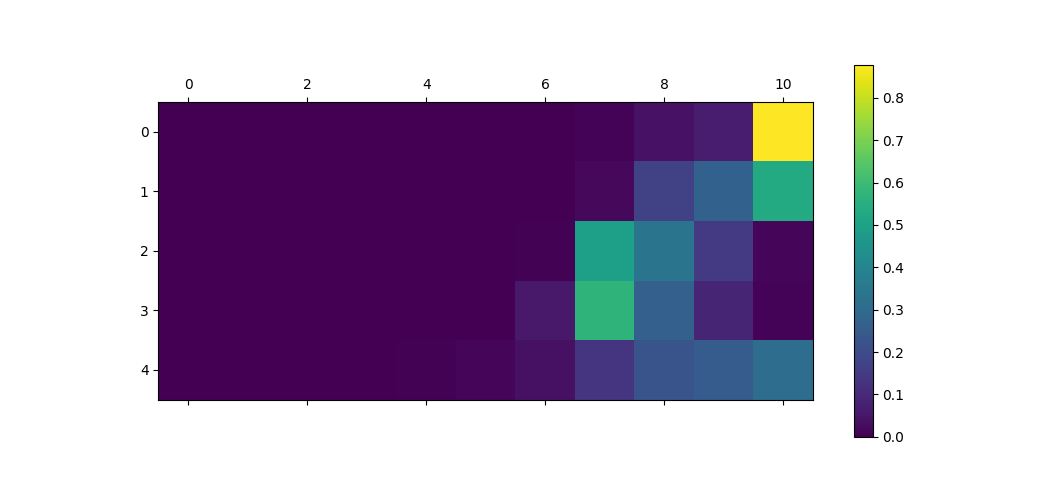}
\includegraphics[height=3.5cm,width=4.31cm]{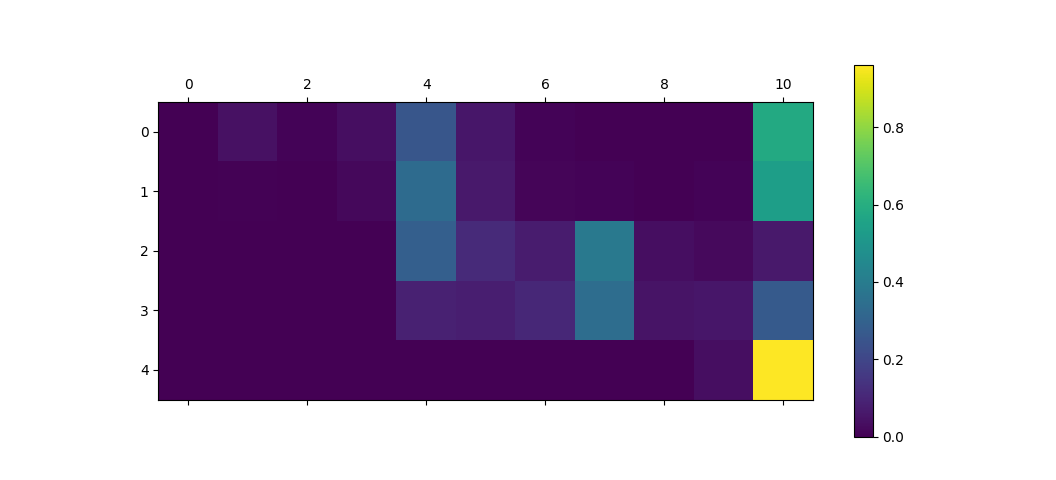}
\includegraphics[height=3.5cm,width=4.31cm]{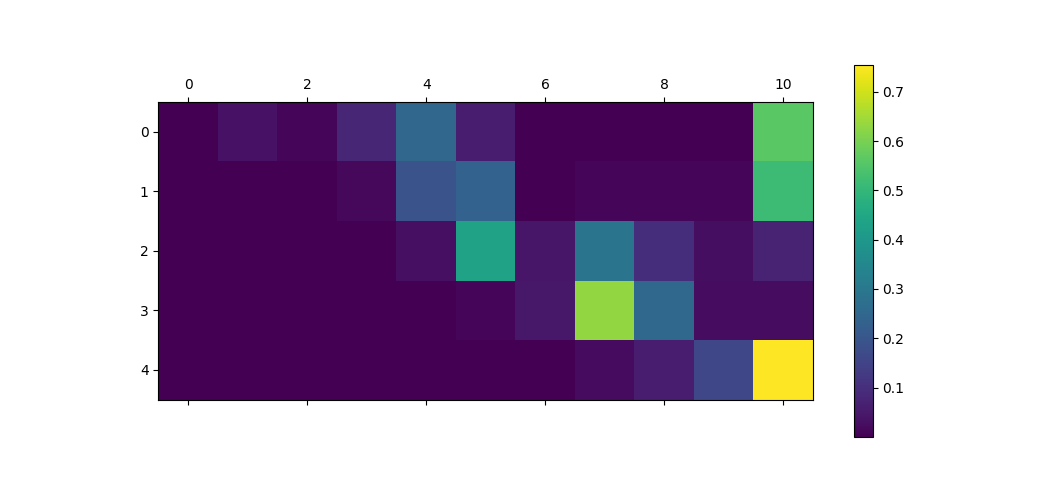}
\caption{Distribution of attention weights learned by DistilBERT$_6$ ALP-KD on CoLA (left), RTE (middle), and MRPC (right). x-axis and y-axis are the teacher and student layer index respectively.}
\label{fig:apl_atn}

\end{figure*}

\subsection{Impact of Random Layer Selection}

To evaluate the impact of random layer selection on the performance of RAIL-KD compared to the other baselines, we report the standard deviation of the DistilBERT$_6$ student models on the three smallest GLUE tasks, which are known to have the highest variance in Table~\ref{tab:std_bert12}. 

Figures shows that RAIL-KD variance is at the same scale compared with PKD and ALP-KD on CoLA and MRPC, and even lower on RTE. This is a strong indicator that the gains from RAIL-KD are not due to random layer selection. 

\begin{table}[!thp]
    \centering
    \begin{tabular}{l|lll|l}
    \toprule
         & \bf CoLA & \bf RTE & \bf  MRPC & \bf  Avg.\\ 
        \midrule
        PKD & $\pm$0.14 & $\pm$1.50 & $\pm$0.24 & $\pm$0.63\\ 
        ALP-KD & $\pm$0.95 & $\pm$1.30 & $\pm$0.70 &$\pm$0.98\\ 
        
        RAIL-KD$^l$ & $\pm$0.49 & $\pm$0.40 & $\pm$0.25 & $\pm$0.38 \\ 
        RAIL-KD$^c$ & $\pm$0.51 & $\pm$0.23 & $\pm$0.40 & $\pm$0.38 \\ \bottomrule
    \end{tabular}
    \caption{Standards deviation (5 runs) of DistilBERT$_6$ ILD models on the smallest three GLUE datasets. We also report the unweighted average on the 3 tasks.}
\label{tab:std_bert12}
\end{table}

\subsection{Out-of-Distribution Test}

We further validate the generalization ability of student models by measuring their robustness to in-domain and out-of-domain evaluation. We do so by evaluating models fine-tuned on MLI, QQP and SST-2 and then evaluated on SciTail, PAWS, and IMDB respectively. These datasets contains counterexamples to biases found in the training data~\cite{mccoy2019right,schuster2019towards,clark2019don}. Performances of \bert{$_{12}$}/Roberta$_{24}$ teacher and DistilBERT$_6$/DistilRoBERTa$_6$ student variants are reported in Table~\ref{tab:ood}. Also, we compute the unweighted average score of the three tasks.

First, we notice high variability in models rank and some inconsistencies in performances across tasks when compared with in-domain results. This was also reported in prior works on out-of-domain training and evaluation~\cite{clark2019don,mahabadi2020end,utama2020towards,sanh2020learning}. Still, RAIL-KD$^{l/c}$ clearly outperforms all baselines across tasks. Surprisingly, we observe that PKD and ALP-KD perform poorly (on all three tasks) compared to the Vanilla KD baseline. 

\begin{table}[!htb]
\centering
		\resizebox{\columnwidth}{!}{	
\begin{tabular}{l|ccc|c}

\toprule
\textbf{Model} & \textbf{SciTail} & \textbf{PAWS} & \textbf{IMDB} & \textbf{Avg.}\\
\midrule

Teacher & 70.3/82.7 & 43.3/43.3 & 84.6/88.9 & 66.0/71.6\\ 
\hdashline
w/o KD & 68.7/74.9 & 36.5/34.7 & 81.3/85.8 & 62.2/65.1\\
Vanilla KD & 68.6/76.1 & 42.2/36.6 & 82.0/86.1 & 64.3/66.3\\
PKD & 68.0/74.8 & 39.9/36.5 & 80.9/85.4  & 62.9/65.6\\
ALP-KD & 66.9/74.7 & 40.7/35.7 & 78.7/82.8  & 62.1/64.4\\

\hdashline
RAIL-KD$^l$& 68.6/\textbf{76.6} & 39.0/\textbf{36.9} & 83.2/\textbf{87.3}  & 63.6/\textbf{67.0}\\

RAIL-KD$^c$ & \textbf{68.7}/75.6 &\textbf{43.7}/36.2 & \textbf{85.0}/85.9  & \textbf{65.8}/65.9\\

\bottomrule

\end{tabular}
}
\caption{Out-of-domain performances of models trained on MNLI, QQP, SST-2 and evaluated on SciTail, PAWS, and IMDB respectively. \bert{$_{12}$}/Roberta$_{24}$ and DistilBERT$_6$/DistilRoBERTa$_6$ are used as backbone for the teacher and students respectively. For each setting, We report the unweighted average score on the 3 tasks.}
\label{tab:ood}

\end{table}

Interestingly, we observe that RAIL-KD$^l$ performs consistently better (1.1\% on average) than RAIL-KD$^c$ on Roberta$_{24}$ compression, while RAIL-KD$^c$ perform better (1.1\% on average) on \bert{$_{12}$}. These results suggest that layer-wise distillation approach is more effective than concatenated distillation when we have a large capacity gap (layer number) between the teacher and the student, and vice versa.

\section{Analysis}

We run extensive analysis to better understand why  RAIL-KD performs better than the other baselines. We visualize the layer-wise cosine similarity between the intermediate representations of the teacher and the student networks. Figure~\ref{fig:plot} shows the cosine similarity score between three intermediate layer representations of \bert{$_{12}$} teacher (i.e. layers 2, 4 and 6) and the first three layer representations of the student for PKD, ALP-KD, RAIL-KD$^{l/c}$ students on 100 samples randomly selected from the SST-2 dataset. Due to space constraints, we only plot the scores for the first three layers of the  student model. The similar trend are seen from the other layers.

We found that RAIL-KD allows the student to mimic teacher layers similar to PKD and much better than ALP-KD, despite that the mapping scheme varies at each epoch. Moreover, we observe that ALP-KD method  gives less similarity scores in the upper intermediate layers. PKD gives lower similarity scores in the lower layers while improving in the upper layers. In contrast, our approach gives more stable similarity scores for all layers and getting closer to the teacher representation in the upper layers. 

We further investigate the attention weights learned by ALP-KD, and find out that they mostly focus on few layers (sparse attention). Figure~\ref{fig:apl_atn} illustrates the distribution of weights, averaged on all training samples of DistilBERT$_6$ ALP-KD studnet on CoLA (left), RTE (middle), and MRPC (right)~\footnote{Similar trends found on other datasets.}. The figure clearly shows (light colors) that most of ALP weights are concentrated on top layers of the teacher. For instance, layers 1,2,5 of the three students mostly attend to the last layer of \bert{$_{12}$}. This may be an indicator that ALP-KD overfits to the information driven from last layers. In contrast, the randomness in layer selection of RAIL-KD ensures a \textit{uniform focus} on teacher layers. This may explain the poor performance of ALP-KD on out-of-domain evaluation compared with RAIL-KD.

\section{Conclusion and Future Work}
We introduced a novel, simple, and efficient intermediate layer KD approach that outperforms the conventional approaches with  performance improvement and efficient training time. RAIL-KD selects random intermediate layers from the teacher equal to the number of intermediate layers of the student model. The selected intermediate layers are then sorted to distill their representations into the student model. RAIL-KD yields better regularization, which helps in performance improvement. Furthermore, our approach shows better performance for larger model distillation with faster training time, which opens up an avenue to investigate our approach for a super-large model such as GPT-2~\citep{GPT} distillation using intermediate layers, as well as to improve robustness and generalization on a wider range of NLU tasks~\cite{ghaddar2021context,ghaddar-etal-2021-end}.

\section*{Acknowledgments}
We thank Mindspore\footnote{\url{https://www.mindspore.cn/}} for the partial support of this work, which is a new deep learning computing framework.

\normalem
\bibliography{custom}
\bibliographystyle{acl_natbib}

\end{document}